\documentclass{article}
\usepackage{graphicx} 
\usepackage{authblk}

\author[1]{Alon Halevy}
\author[1]{Jane Dwivedi-Yu}
\affil[1]{Meta AI}

\title{Learnings from Data Integration for Augmented Language Models}
\date{March 2023}

\begin{document}

\maketitle

\begin{abstract}
    One of the limitations of large language models is that they do not have access to up-to-date, proprietary or personal data. As a result, there are multiple efforts to extend language models with techniques for accessing external data. In that sense, LLMs share the vision of data integration systems whose goal is to provide seamless access to a large collection of heterogeneous data sources. While the details and the techniques of LLMs differ greatly from those of data integration, this paper shows that some of the lessons learned from research on data integration can elucidate the research path we are conducting today on language models.  
\end{abstract}

\section{Introduction}

Large language models (LLMs) have shown impressive performance recently and their chat-based interfaces demonstrate the promise of ushering in a new and powerful paradigm for information access, one that is 
 a huge step up from today's search engines. To realize this new paradigm, we need to overcome some of the known limitations of LLMs, including access to up-to-date information, and the ability to point to the source of their answers as a method of mitigating their tendency to hallucinate. 
To that end,  providing LLMs access to external sources (a.k.a.\ tools or plugins) has become a very active area of research with rapid progress~\cite{komeili-etal-2022-internet, thoppilan2022lamda,lazaridou2022internet,shuster2022blenderbot,yao2022react,nakano2021webgpt,cobbe2021training,gao2022pal, schick2023toolformer}. In essence, the vision is that the LLMs will be a mediator to many sources of high-quality information and reasoning capabilities.

Data integration systems (a.k.a.\ as mediator systems or information gathering agents) have a similar goal of providing seamless access to a multitude of data sources. They were inspired by the need of enterprises to easily query the many disparate databases they owned, large-scale collaborative science projects that had to reconcile data from multiple research labs, and later on, the dream of seamlessly querying any available data on the WWW. Unlike LLMs whose knowledge is broad and can speak of anything from ethics to celebrity gossip, data integration systems were built to answer queries in a particular domain of interest. 

The point of this short article is {\em not} to argue that data integration systems solved the challenges we are facing today. Quite the contrary, from a data integration perspective, the latest developments are a dream come true. However, an important contribution of the data integration literature has been to map out the problem space of integrating a large number of data sources and address them one by one using the techniques available at the time (think, mathematical logic).    However, we argue that the distinctions identified in the problem space are equally relevant to today's challenges and highlighting them might add clarity as we pursue our current research.  

\section{Data integration vs.\ augmented LLMs}
The goal of a data integration system is to provide seamless access to a (possibly large) set of data sources. The definition of what is considered a data source is rather broad. It can be a database, a knowledge base, or any store of information behind an API that is able to receive queries and return answers from the source.
It is also important that we should be able to describe the contents of the data source with a schema (even if it's not the actual schema in which the data is stored).  We briefly describe the components of a data integration system and then map them to the terminology of today's language models. For a book-length description of data integration, see~\cite{DBLP:books/daglib/0029346}.

A user poses queries to the data integration system through a {\em mediated schema}, which is the set of terms that the system models, such as classes of objects,   their attributes and relationships between objects. For example, an application about movies would model the classes {\sf Movies}, {\sf Actors} and {\sf Directors}, the attributes {\tt title},   {\tt year} and {\tt review}, and the relationships {\tt actedIn} and {\tt directedBy}. The data sources store different slices of the domain data in their own schema, often referred to as source schema. For example, one data source $S1$ may store movies that were produced after 1960 and another, $S2$, may store all French movies along with their reviews. The connection between the mediated schema and the source schemas is done by a set of schema mappings that are expressions in a specialized logic. 

To answer a query $Q$, the data integration system needs to first reformulate $Q$ into a query $Q'$ that refers to the source schemas and can actually be executed. The  reformulator is a specialized reasoning engine that takes the query $Q$ and the semantic mappings as input and produces the query $Q'$, and is in some sense the heart of the system. The query $Q'$ may be of the form {\em take the union of  source S1 and source S2 and then filter to year$\geq$2000}. The query $Q'$ is executed by making multiple calls to the data sources and perhaps supplementing these calls with additional processing inside the data integration system itself.   

In today's setting, the data sources  still need to expose an API.  The key difference from the data integration setting is that the user interacts with the system using natural language and can ask about any topic under the sun, and so there is no mediated schema. This also means that the descriptions of the data sources focus on expressing what kind of information the source contains, rather than providing mappings between sets of symbols.  The key challenge is to {\em learn the mediation process}, i.e., to train the language model to decide which sources are relevant to the query, perform the appropriate reformulation to the source's API, and to combine and reason about the results to produce the final answer.   Another major difference worth noting in the current context is that the language model itself encodes a vast amount of knowledge. This knowledge will often be used to answer questions without going to any external source or to fill in the gaps between the data coming from the sources and user's need.
As we illustrate in the next section, the problem space as laid out in the data integration literature, can be helpful in understanding the nuances of the mediation problem.

\section{The common challenges}  
\label{section:reformulation}
The first distinction in the space of reformulation problems is whether answering the user's question will require accessing a single data source, or whether it will require combining information from multiple sources. Section~\ref{section:single} considers the case of a single source and Section~\ref{section:join} considers the additional complications when we need to chain multiple sources.

\subsection{Answering questions from a single source}
\label{section:single}

Unlike the data integration setting, here we want the model to learn how to mediate. The first challenge is for the system to learn how to discriminate between the contents of the sources. Specifically, if there are multiple sources available to the system, the model needs to figure out which ones are the most relevant to the question it was given. Moreover, it should be the case that  adding or removing sources from the system  should not require major effort, and in particular, should not entail significant retraining of the model. Ideally, adding a source should only require showing examples of what the new source knows. In the data integration literature, discriminating between the contents of sources was achieved by providing more expressive logical descriptions of their contents and the Local-as-View approach to describing sources ensured that sources can be described in isolation from others in the system~(\cite{DBLP:books/daglib/0029346}, Chapter~3). It is worth noting that 
 these challenges already arise in the Toolformer system~\cite{schick2023toolformer}. Adding a new source requires that the model be retrained, and the system struggles when sources have competing contents.

\medskip
\noindent
{\bf Discriminating between sources:}
While sources with broad coverage (e.g., a web search engine and Wikidata) may indeed compete often, we argue that that's a relatively uncommon case. Here too, we draw on the 
data integration literature that teased out the different cases of when sources may compete with each other, and it is instructive to know which cases are being faced when training the model to mediate. 

In the simplest case, if sources cover completely different domains (e.g., a source about recipes and one about stock quotes) then the model should have no problem discriminating between them. The challenge starts when sources are related, and there are a few common cases here:
\begin{itemize}
    \item Different properties of the same entities: for   example, one source may provide the addresses and the menus of restaurants, while the other may provide customer reviews. 
    \item Different entities, same properties: for example, one source may contain information about American universities while another may cover French ones. 
    \item Subsumption: in this case, one source contains a superset of the information of the other. In general, the model can lean on the bigger data source, but there may be reasons for choosing the smaller one, such as better quality or response time. 
    \item Source completeness, authoritativeness: there are other properties of sources that can affect the reformulation. For example, some sources may have complete coverage of all or of a known subset of their domain (e.g., IMDB for movies). Other sources may be considered more authoritative than others and hence preferable.    
\end{itemize}

One might argue that over time the world of information will evolve to having only few data sources and those will have all the information we need. For example, Wolfram Alpha has access to a huge amount of data, as do the Google Knowledge graph or Wikipedia. Even ignoring the fact that under the covers some of these systems integrate data from multiple other sources, human nature is such that there will likely always be multiple sources that consider the same and/or overlapping domains of data.

\medskip
\noindent
{\bf Source capabilities:} Learning the capability of a data source is also critical because the model may need to use other reasoning to compensate. For example, consider the question {\em  what is the second largest country in Central America (by land size)}. Wolfram Alpha  can answer the entirety of that question because it contains the relevant data and the computational abilities. However, if all we have is access to WikiData that has no reasoning power, then we need to (1) reformulate the question to ask for landmass of countries in Central America, and then (2) compute the second largest. 

\medskip
\noindent
{\bf Correctly accessing the source API:}
The issue of posing the right query to the source's API can be a thorny one. In Toolformer's case, the system initially focused on sources that have a natural language interface so the API was relatively simple.  Wolfram Alpha can also take natural language as input. However, if the source is an actual database and an API call needs to encode an SQL-like query, this may be more tricky to get right. It is conceivable that LLMs' ability to generate code may be enough to address this challenge (since queries are essentially code). 

The discussions on standards for LLM plugins are nascent, but one of the issues to note is that  the API be able to return different kinds of {\em don't know} answers, such as: (1) the query is ill formed, (2) the query is well formed but nonsensical (e.g., who is the mother of France), (2) the answer is empty (children of Oprah), or (3) the answer may be incomplete or wrong.  

\subsection{Chaining multiple sources}
\label{section:join}

When a question cannot be answered using a single source, we need a chain of inferences that combines multiple sources.  As noted earlier, since the LLM itself has vast knowledge, it can be considered a source as well. Research on chain of reasoning is already ongoing in the community \cite{wei2022chain}, and we expect the chains that a language model will exhibit to be much more complex than the ones that data integration systems were capable of. However, it is still worth noting a few types of  chains that occur frequently and some other issues that arise when sources are composed. 

\medskip
\noindent
{\bf Chains of reasoning:} The common patterns of reasoning that database and data-integration systems pursue are the following.
\begin{itemize}
    \item Join: the result of one source is fed to another (or filtered by another). For example, the query {\em in which years did a  European country win  the world cup?} would require a source that knew which continent countries are on and a source about the world cup.

    \item Union: we obtain similar results from different sites and take their union. For example, {\em find all conferences in Boston this year} may require going to a few recipe sites. 
    
\end{itemize}

The above operations can be composed to answer more complicated questions, and in addition the question may require an aggregation step (e.g., {\em how many European countries won the world cup}). An important point about all of these chains of reasoning is that they often require reasoning about {\em sets of answers}, something that language models still struggle with considerably. 

\medskip
\noindent
{\bf Constraints on ordering sources:} When chaining multiple sources there may be constraints on their possible ordering (known as binding-pattern restrictions~\cite{DBLP:conf/pods/RajaramanSU95}).  For example, for the question {\em which MIT professors have an H-Index greater than 100},  the chaining needs to start from the source that provides the list of MIT professors, and then query the site that provides the H-Index, since the latter does not support the query for all H-indexes of everybody. In some cases, including the former, ordering the sources is an important optimization consideration (as is join ordering in SQL systems). For example, to answer {\em which French nationals won the Turing Award}, it is much more efficient to start from Turing Award winners and then try to find out which ones are French.   The efficiency and possible monetary cost of a  tool can also be a factor. For instance, if we  need to add two numbers, a calculator will suffice, whereas   the much more powerful Wolfram Alpha may be an overkill and slower. 

\medskip
\noindent
{\bf Domain-level vs.\ source-level reasoning:} As we train the LLM to reason about external sources, it's interesting to see whether what it is learning is at the level of language or at the level of individual sources. For example, if we swapped one recipe source with another (that has similar content characteristics), would most of what the LLM learn transfer to the new source, or is its knowledge somehow tied to the previous one.

\medskip
\noindent
{\bf Entity matching:} One of the thorniest issues in data integration and likely to apply in this context as well is matching entities across multiple sources. One cannot combine data from multiple sources unless it can be determined that the data speak of the same entities (e.g., refer to the same person, organization or product). The topic of entity matching has received significant attention (and ironically, has been known by several different terms, such as reference reconciliation, tuple matching and others), and it's unclear whether the state-of-the-art results will transfer into the LLM setting.

\section{Conclusions}
Data integration systems and LLMs accessing external data share a common vision of providing seamless access to huge bodies of diverse data. Clearly, the context of LLMs is much more powerful, with more diverse data sources and much more interesting queries. The techniques from the data integration literature will not help us figure out {\em how} to train an LLM for reformulation. However, as we demonstrated above, it will shine light on {\em what} the LLM should learn, and in turn, this may contribute to the research agenda. 

\bibliographystyle{IEEEtran}
\bibliography{custom}

\end{document}